\documentclass{article}
\usepackage{cite}
\usepackage{amsmath,amssymb,amsfonts}
\usepackage{hyperref}
\usepackage{algorithmic}
\usepackage{graphicx}
\usepackage{textcomp}
\usepackage{xcolor}
\def\BibTeX{{\rm B\kern-.05em{\sc i\kern-.025em b}\kern-.08em
    T\kern-.1667em\lower.7ex\hbox{E}\kern-.125emX}}
\begin{document}

\title{Streetify: Using Street View Imagery And Deep Learning For Urban Streets Development}

\author{
Fahad Alhasoun \\ \textit{Massachusetts Institute of Technology} \\ fha@mit.edu
\and 
Marta Gonz\'alez \\ \textit{University of California, Berkeley} \\ martag@berkeley.edu
}
\maketitle
\begin{abstract}

The classification of streets on road networks has been focused on the vehicular transportational features of streets such as arterials, major roads, minor roads and so forth based on their transportational use. City authorities on the other hand have been shifting to more urban inclusive planning of streets, encompassing the side use of a street combined with the transportational features of a street. In such classification schemes, streets are labeled for example as commercial throughway, residential neighborhood, park etc. This modern approach to urban planning has been adopted by major cities such as the city of San Francisco, the states of Florida and Pennsylvania among many others. Currently, the process of labeling streets according to their \textit{contexts} is manual and hence is tedious and time consuming. In this paper, we propose an approach to collect and label imagery data then deploy advancements in computer vision towards modern urban planning. We collect and label street imagery then train deep convolutional neural networks (CNN) to perform the classification of \textit{street context}. We show that CNN models can perform well achieving accuracies in the 81\% to 87\%, we then visualize samples from the embedding space of streets using the t-SNE method and apply class activation mapping methods to interpret the features in street imagery contributing to output classification from a model.
\end{abstract}


\section{Literature}

There have been several interesting application of machine learning and particularly deep learning on imagery data in the domain of urban computing and remote sensing. Researchers from the domain of remote sensing have been focusing on developing models of inferring land use and land cover from satellite imagery\cite{albert2017using,Buslaev_2018_CVPR_Workshops,Zhou_2018_CVPR_Workshops,Hamaguchi_2018_CVPR_Workshops,Aich_2018_CVPR_Workshops,albert2017modeling}. Recently, researchers organized the Deepglobe challenge as part of the CVPR conference, the challenge focused on common challenges in remote sensing including road extraction, building detection and land cover classification \cite{demir2018deepglobe}. Researchers produced a myriad of approaches and deep learning techniques to address the challenging tasks\cite{demir2018deepglobe,zhou2018d,hamaguchi2018building,sun2018stacked,kuo2018deep}. Furthermore, researchers used street view imagery in congruence with satellite imagery to enhance the process of digitizing maps.  Cao et al. used an integrative approach of using satellite imagery and street view to infer land use then classify buildings and points of interest (POIs)\cite{cao2018integrating}.


Using street view imagery researchers showed how it can be a predictor to several urban socio-economic measures in cities. Gebru et al. developed an object detection model to extract cars from street view images, the paper then proceed by doing an image classification of the make, model and age of cars seen in a neighborhood area\cite{gebru2017using}. Then a lookup for the prices of the cars seen in images is performed against a database of their expected prices. The paper shows statistically significant correlations between demographics and the prices of cars residing in an area. They tested the strength of correlation of the metric of extracted car prices against several demographical features of urban areas, namely the U.S. Census and presidential voting data. Naik et al. used street view imagery to develop models that can predict the perceived safety of a street. They trained models against perceived street safety collected through surveying 7000 participant to input their perceived safety score given a street view image\cite{naik2014streetscore}. Naik et al. extended the work and developed a computer vision model to measure changes in the physical appearances of neighborhoods from street-level imagery across time. They studied the correlation of the magnitude of change to neighborhood demographical characteristics. They identified metrics that can predict neighborhood physical change, they found that education level and population density are strong predictors in magnitude of urban physical change of an area\cite{naik2017computer}. Kidzinski et al. used street view imagery of houses and apartment complexes to train a model against car insurance data. The paper shows predictive strength in visual features of houses towards predicting risk of car accidents. Given the address of insurance beneficiary, the model inputs street imagery of their home address and can outputs prediction of car accidents risk of an individual\cite{kita2019google}. The visual features in street view images are very rich enabling many applications of urban computing. We explore its potentials for urban planning towards better classifying streets in a city.

\section{Data}
The city of San Francisco developed a manual process by which street contexts are determined. In this section, we first discuss the manual process of street context classification in urban planning developed by the city of San Francisco \cite{SF_BSP}. Then, we proceed by discussing how we sample labeled street imagery data for the purpose of training a deep convolutional neural network to perform the task of street context classification.
\begin{figure}[!ht]
	\begin{center}
		\includegraphics[width=\linewidth]{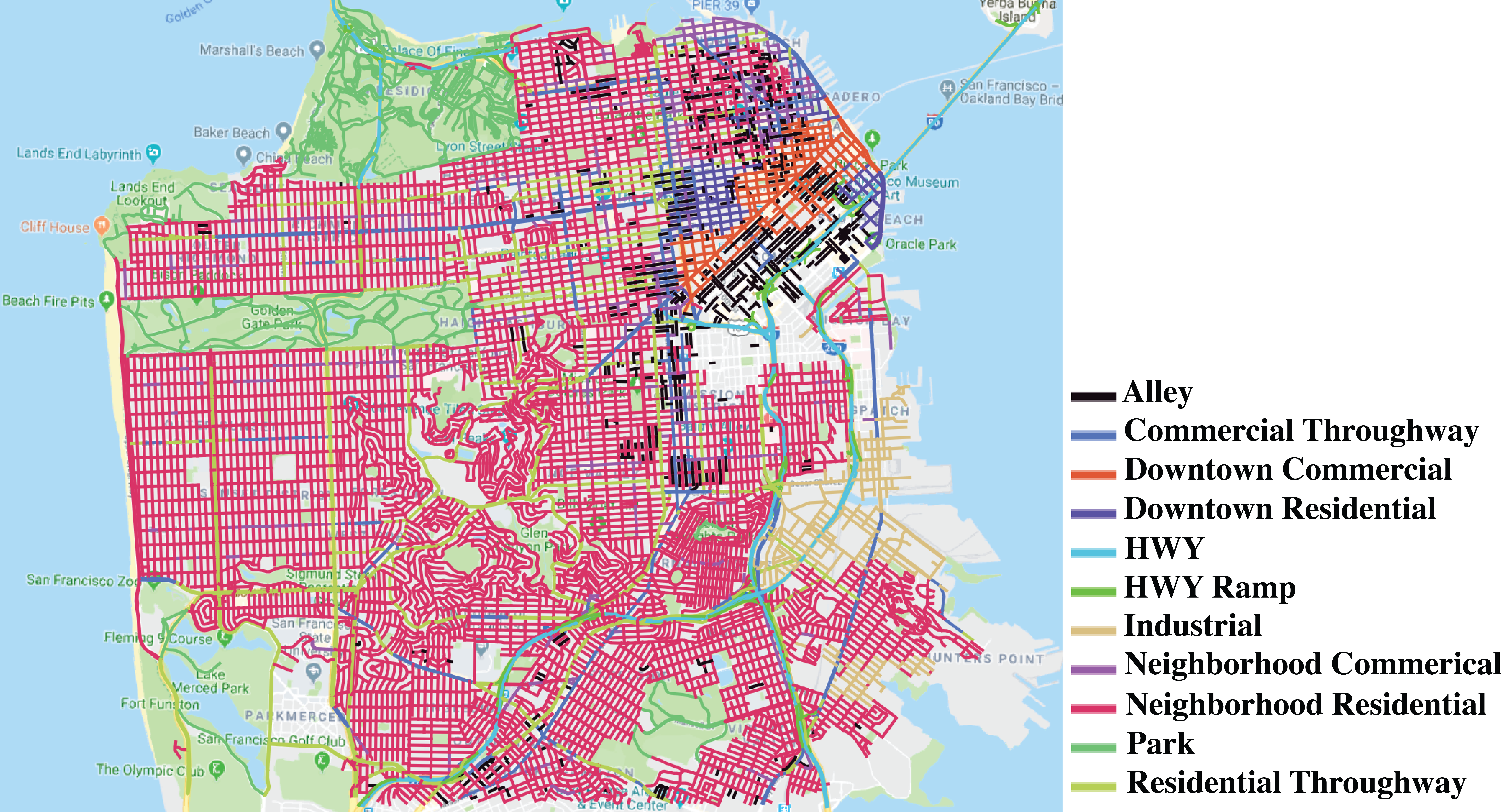}
		\caption{The road network of San Francisco with labeled street contexts.}
		\label{fig:SF_labels}
	\end{center}
\end{figure}

\subsection{The manual street context classification process}
The city of San Francisco developed a manual process by which street contexts are determined\cite{SF_BSP}. The inputs of this manual process of street context classification are the shapefiles of the unlabeled streets in a city and the use of parcels on the sides of the streets (i.e. commercial, residential...etc). We first collect the streets shapefile for the city of interest, the data is usually made available by city councils. The shapefile data for the streets of Boston and San Francisco were graciously provided by the city councils (San Francisco streets were already labeled by the city council). The city also shared the manual process used by their urban planners to perform the classifications of street contexts. We then followed the manual process to label streets in Boston. The labeling scheme results in 11 classes for the city of San Francisco and 10 classes for the city of Boston. This is due to the reason that there is little to no presence of the \textit{Downtown Residential} class in the downtown of Boston. 
  \begin{figure}[!h]
 	\begin{center}
 		\includegraphics[width=\linewidth]{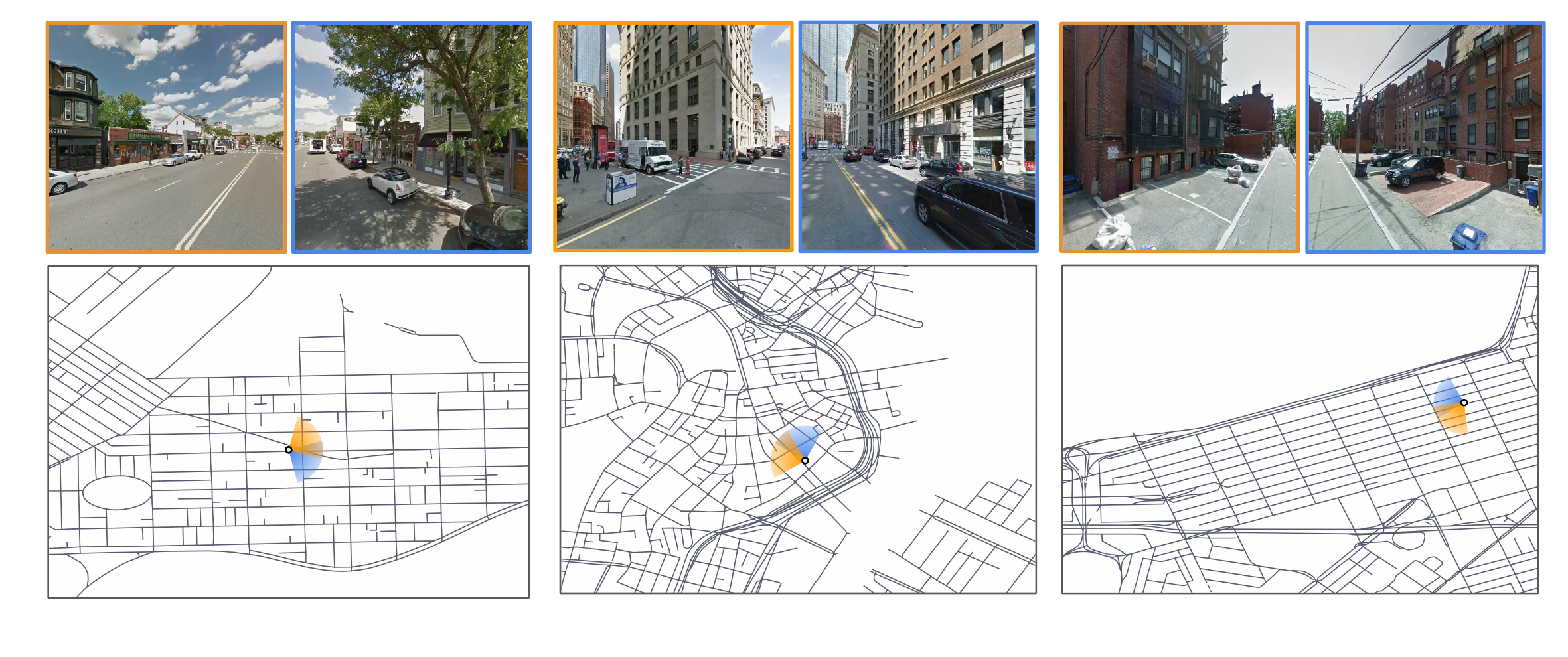}
 		\caption{ (From the left) Commercial Throughway, Downtown Commercial and Alley in Boston.}
 		\label{fig:threeSamples}
 	\end{center}
 \end{figure}

The streets are labeled based on conditions pertaining to transportational functionality and the use of land on the sides. The San Francisco classification scheme developed by the city's urban planners is a multistage scheme for classification of streets that can be summarized in the following steps: 
\begin{enumerate}
\item Determine the side use context: Streets are distinguished by their side use using parcel information including side contexts of \textit{commercial} and \textit{residential}. 
\item Determine the transportation context: we then assigning labels that pertain to transportational features of the street. This includes  \textit{throughway}, \textit{highway} and \textit{highway ramps}. Streets with lower flows are labeled \textit{dowtown} and \textit{neighborhood} for their transportation context. 
\item Identify special conditions: certain streets have special classifications including \textit{alleys}, \textit{parks}, \textit{industrial}.  
\end{enumerate}
\subsection{Sampling labeled street imagery:}
   Figure \ref{fig:SF_labels} shows the 11 classes for the city of San Francisco. Streets of the class \textit{Neighborhood Residential} constitute the majority of the streets. The process of classification for the city of SF was conducted by the city council. We followed the same process to label subset of streets in Boston for this study.


Once a set of streets in a city is labeled according to the manual process of street context classification discussed in the previous subsection, we then can use the shapefiles of streets to sample labeled street view imagery. First, we randomly sample a street segment from the subset of labeled street segments without replacement. Then, we sample a random lat/lon from the sampled street segment. We then proceed to collect images that capture  sides of the street as well as the road ahead.  Images provided by the Google Street View API cover an angle of approximately 90 degrees.  To cover the sides of the street while maintaining view on the road, we collect two Images one tilted towards the right of the direction of traffic on the street and the other tilted towards the left of the direction of traffic on the street. The pair of images that we sample are labeled with the street context label as per the shapefile. 

 The Google Street View images API \cite{streetViewApi} provides street view imagery as a service. The service provides a free quota for the use of the API that we utilize for the purpose of this study, the free quota is renewed every month. The API takes the lat/lon of a location as well as an angle of view and returns images of varying sizes and maximum size of 640x640 pixels. Figure \ref{fig:threeSamples} shows sample points on the street road network and the corresponding views from the street view API. In orange frame are the images taken with an angle towards the left and in blue frame are images that are taken towards the right of the direction of the street. From left to right of the figure we have a Commercial Throughway, a Downtown Commercial and an Alley from Boston. Each sample image has a map below it showing the location where the lat/lon and view angle of the images.

\section{Modeling}
In this section we discuss the conceptual framework of the process and the architectures of the deep Convolutional Neural Networks (CNN) in our framework.
\subsection{The general framework}

\begin{figure}[!h]
	\advance\leftskip-4cm
	\begin{center}
		\includegraphics[width=\linewidth]{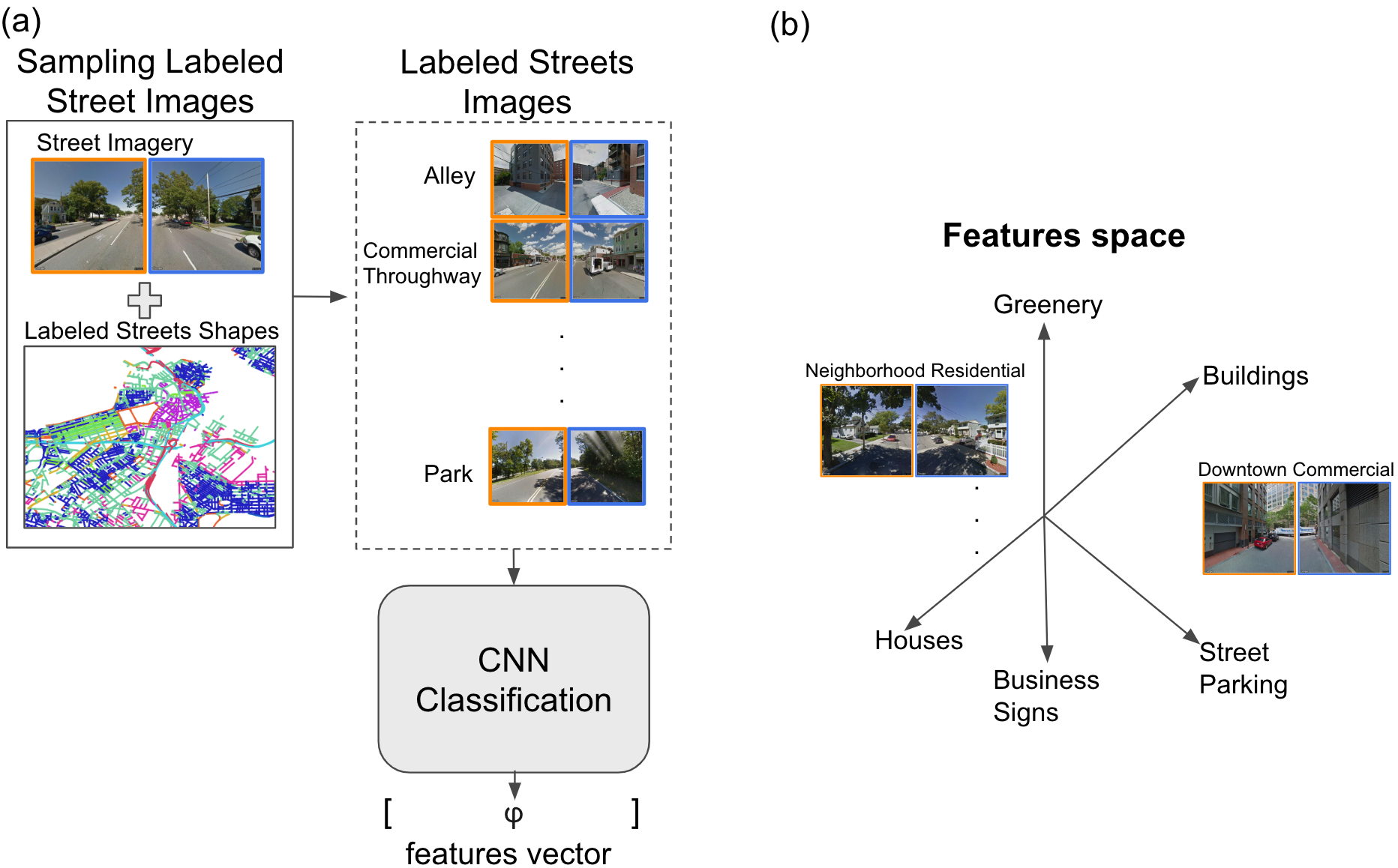}
		\caption{The general framework architecture (a) The pipeline of labeling and sampling street view images then training a CNN model. (b) The continuous embedding space of street contexts.
}
		\label{fig:25}
	\end{center}
\end{figure}

Street context classification incorporates the side use of streets and land use of its sides in addition to the transportational attributes of a street. Side use of streets is influenced by the cultural and socio-economic functions the street servers (which is a subjective judgment by experts). Figure \ref{fig:25}-a illustrates the framework of street context classification of a city. The framework sampling street view imagery from labeled streets shapefile. The CNN model is trained to perform the classification. The CNN outputs a feature map in the embedding space of street contexts.  

We take the view that street context classes are just a useful discretization of a more continuous spectrum of patterns in the organization of fabric of streets in an urban setting. This viewpoint is illustrated in Figure \ref{fig:25}-b while some attributes (e.g., amount of built structures or vegetation) are directly interpretable, some others may not be. Nevertheless, these patterns influence, and are influenced by, socio-economic factors (e.g., economic activity), and dynamic human behavior (e.g. mobility, parking occupancy). We see the work on cheaply curating a large-scale street view classification dataset and comparing streets using deep representations that this paper puts forth as a necessary first step towards a granular understanding of urban settings in data-poor regions.

\subsection{Convolutional Neural Network (CNN) architectures}

In 2012, Krizhevsky et al. applied a CNN to the Imagenet \cite{krizhevsky2012imagenet}. It was the first time an architecture was more successful than traditional, hand-crafted feature learning on the ImageNet. The AlexNet laid the foundations for the traditional CNN, a convolutional layer followed by an activation function followed by a max pooling operation. Much of the success of deep neural networks has been accredited to these additional layers. The intuition behind their function is that these layers progressively learn more complex features. The first layer learns edges, the second layer learns shapes, the third layer learns objects, and so on. In this paper, we explore various CNN architectures starting with the AlexNet then moving to more recent ones including ResNet and Inception \cite{krizhevsky2012imagenet,szegedy2016rethinking,he2016deep}.

\section{Results and Validation}
In this section, we show the accuracies of the models per city, per model architecture. We show the confusion matrices of the Inception-v3 model on the validation set of Boston and San Francisco. The data was split 80\% for training/testing and 20\% for validation. 

\subsection{Accuracy}
 Table I shows the accuracies of the different architectures trained on labeled images from the city of Boston and San Francisco. The AlexNet architecture achieves the lowest accuracy on the validation set for both cities. The Inception-v3 model achieves the highest accuracy on our validation set of both cities. For Boston, AlexNet has an accuracy of 83.16\% and Inception-v3 has an accuracy of 87.79\%. For San Francisco, AlexNet has an accuracy of 81.69\% and Inception-v3 has accuracy of 84.17\%. We notice a drop in accuracy between Boston and San Francisco for the same model architectures generally. This is attributed to the number of classes of streets where there are 11 classes in San Francisco and 10 classes in Boston, the context of \textit{Downtown Residential} is absent for the city of Boston as discussed earlier.
\begin{center}
\begin{table}[ht!]
    \centering
    	\caption{The accuracy of CNN architectures on validation dataset for Boston and San Francisco.}
	\begin{tabular}{|l|c|c|}
		\hline
		CNN/city             & Boston & San Francisco   \\ \hline
		ResNet18                  &  85.64\%    &  81.72\%                 \\ \hline
		ResNet34  &   85.45\%     &            82.02\%     \\ \hline
		ResNet50  &  85.64\%     &               82.71\%  \\ \hline
		AlexNet    &  83.16\%     &         81.69\%    \\ \hline
		Inception-v3	 &  87.79\% &               84.17\%  \\ \hline
	\end{tabular}
 \label{tab:accuraciesTable}
\end{table}
\end{center}

\begin{figure*}[!h]
	\advance\leftskip-4cm
	\begin{center}
		\includegraphics[width=0.8\linewidth]{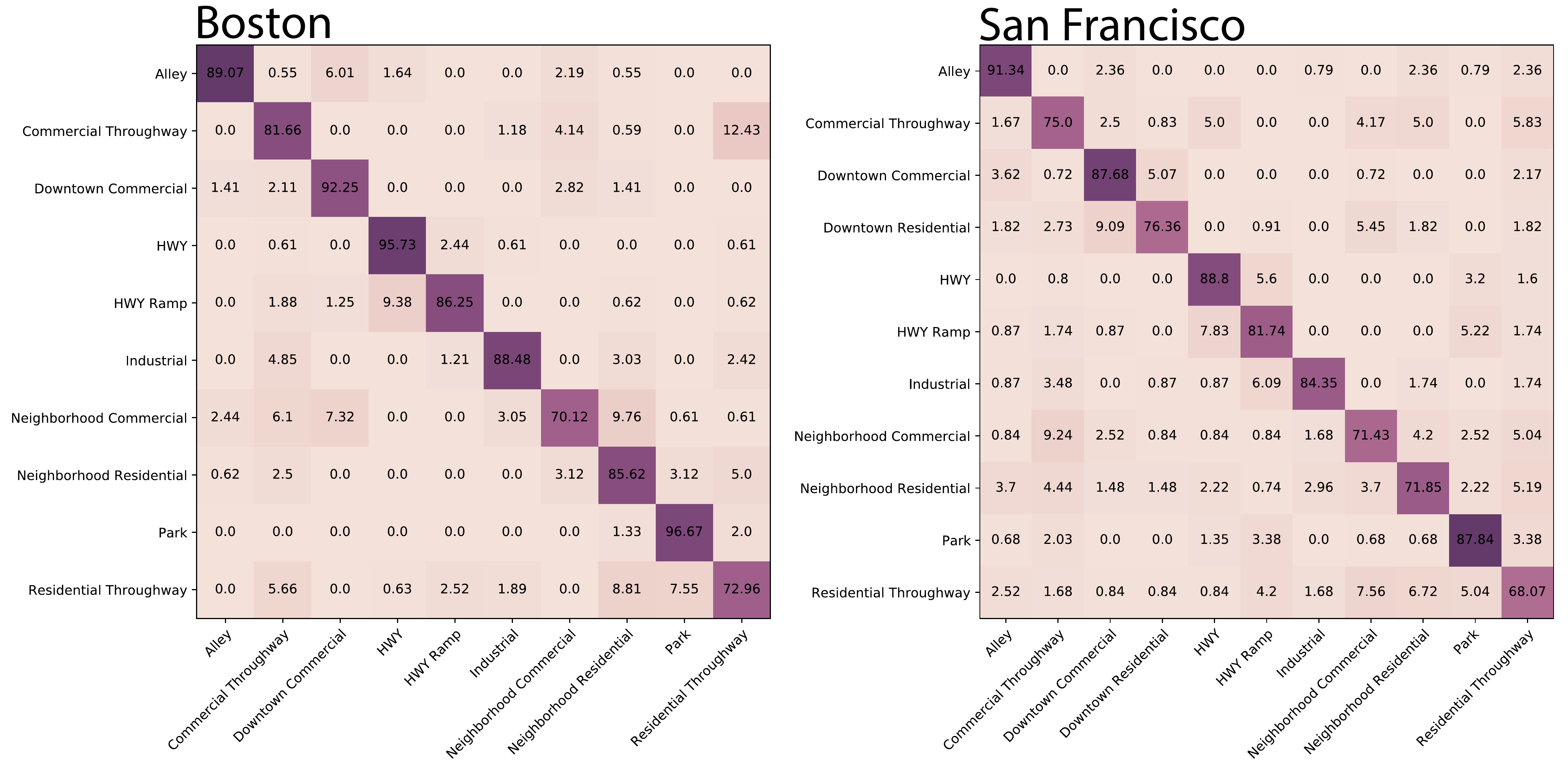}
		\caption{ The confusion matrix for the Inception-v3 model trained on Boston and San Francisco from left to right. The vertical axis has the true labels and the horizontal axis has the predicted labels by the model.}
		\label{fig:cmatresnet}
	\end{center}
\end{figure*}
\subsection{Visualizing the embedding space of street contexts}
\begin{figure*}[!h]
	\advance\leftskip-4cm
	\begin{center}
		\includegraphics[width=158mm]{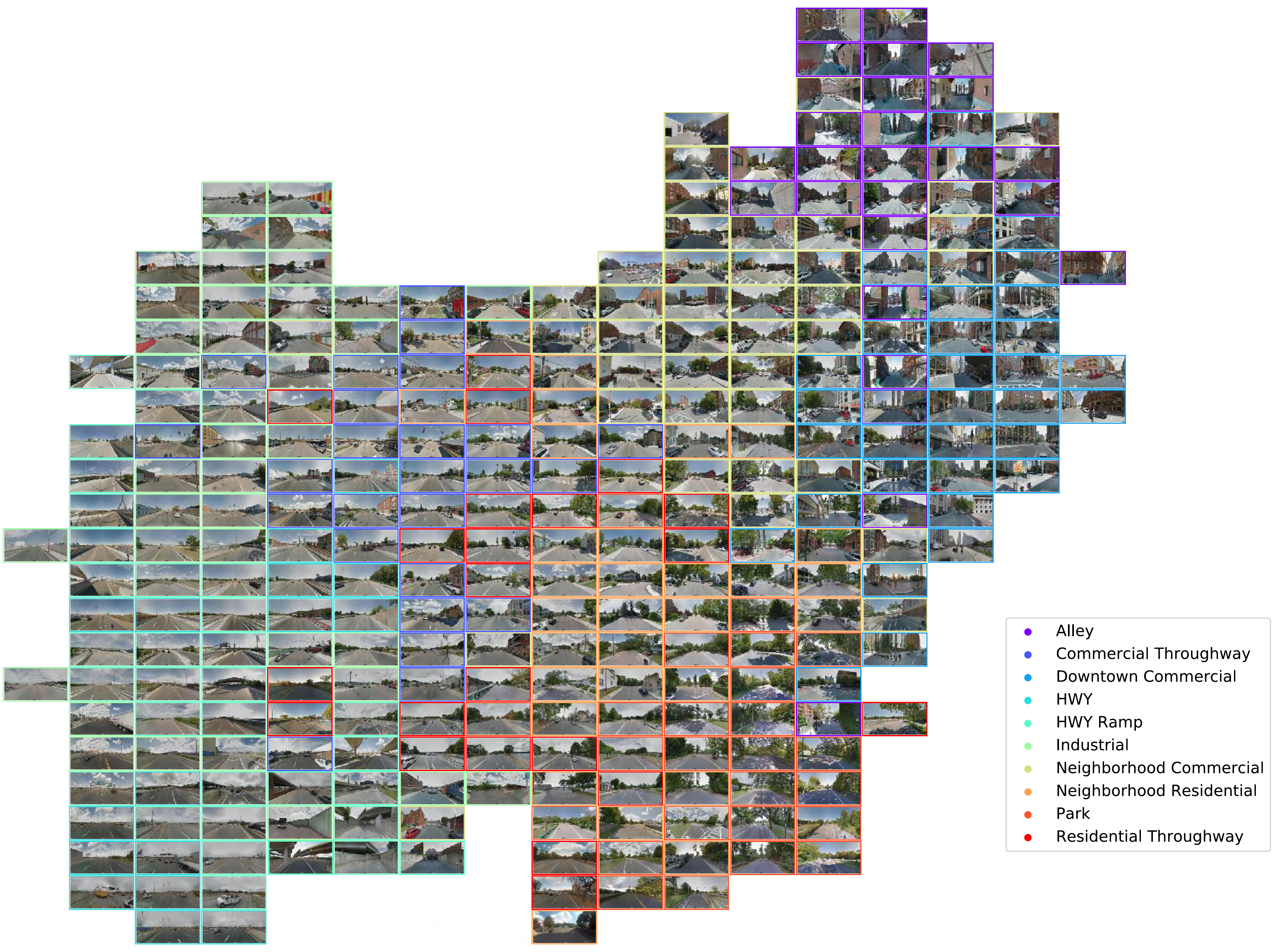}
		\caption{t-SNE visualization of the feature vectors on the 2d space for sample of images from the city of Boston}
		\label{fig:tsne}
	\end{center}
\end{figure*}

Figure \ref{fig:tsne} visualizes the t-SNE projection of the feature vectors for each image in a sample from the training dataset. The feature vectors are the output values on the before last layer on the neural network (in our case we used the AlexNet model architecture). The feature vectors are 4096 dimensional. The t-SNE method help in visualizing the feature vector space by projecting it into a lower dimensional space while preserving neighborhood structure in the original space of the feature vectors\cite{maaten2008visualizing}. Figure \ref{fig:tsne} show the projection of feature vectors for a sample set of street images in Boston onto two dimensions.
The t-SNE visualization shows the neighborhood structure of the highly dimensional spaces of feature vectors. The visualization illustrates the neighboring structure of the high dimensional space on the 2d in our case. The visualization shows the variations in streets in their contexts and geometrical features. Alleys present on the top right side of the plot are narrow and surrounded by red bricked building walls. Street passing through parks contexts are in the lower side of the plot with dense presence of greenery in them. Streets of type Downtown Commercial and Neighborhood Commercial share similar visual attributes to those of Alleys where they are usually surrounded by more buildings that have commercial signs, the Neighborhood Commercial sometimes has more vegetation. They reside on the top right side of the plot showing more of red-bricked buildings on the sides of the streets and little to no vegetation. Neighborhood Residential and Residential Throughway streets are more similar to Parks and have significant presence of greenery in Boston, they reside in the lower right to middle side of the plot. Streets that are highways, highway ramps and industrial contexts reside on the left side of the plot. They are usually wide and have less greenery or presence of buildings on the sides. Commercial Throughways are present closer to middle of the t-SNE visualization where they have some presence of greenery as well as businesses on the sides making them sit between Downtown-like streets streets and Park-like streets.
Figure \ref{fig:cmatresnet} shows the confusion matrices for the Inception-v3 model. The accuracy of the model typically varies by street context. In Boston, Inception-v3 has the highest accuracy for the Park context and lowest for Neighborhood Commercial context. We also notice a few cells where the model confuses streets contexts. For example, the model confuses Highways with Highway Ramps which share similar visual features. The model confuses Residential Throughways, Parks and Neighborhood Residential. They share the visual features of dense greenery on street sides. The model also confuses the Highways and Highway Ramps contexts. Generally, we notice that the cofusion patterns of the model are consistent with our t-SNE visualization where confused classes are usually in close proximity in the embedding space.  

For San Francisco, the model has the highest accuracy for the Alleys and lowest accuracy for classifying the Residential Throughways. The model confuses Downtown Commercial and Downtown Residential. Similar to Boston, the model confuses Highways and Highway Ramps. The mentioned confused classes share similar visual characteristics.

The patterns of confusion between classes in the city of Boston and San Francisco are similar. This is clear in the bottom right corner for the classes of Residential Throughway, Park and Neighborhood Residential. The same holds for Highways and Highway Ramps. In addition to Neighborhood Commercial and Neighborhood Residential. 

\subsection{Class Activation Mapping of street contexts}
To better understand the features which the model is looking for to classify streets. We further investigate the features attributing to the activations on the images using ideas of Class Activation Mapping (CAM) proposed by Zhou et al.\cite{zhou2016learning}. The methodology constructs a heat maps indicative of the features in images that are responsible for activating the predicted class. The heat map is generated by a weighted sum of the last set of convolutional outputs. The weights are of the last layer and corresponding to the outputted class node on the network.


\begin{figure*}[!t]
	\advance\leftskip-4cm
	\begin{center}
		\includegraphics[width=\linewidth]{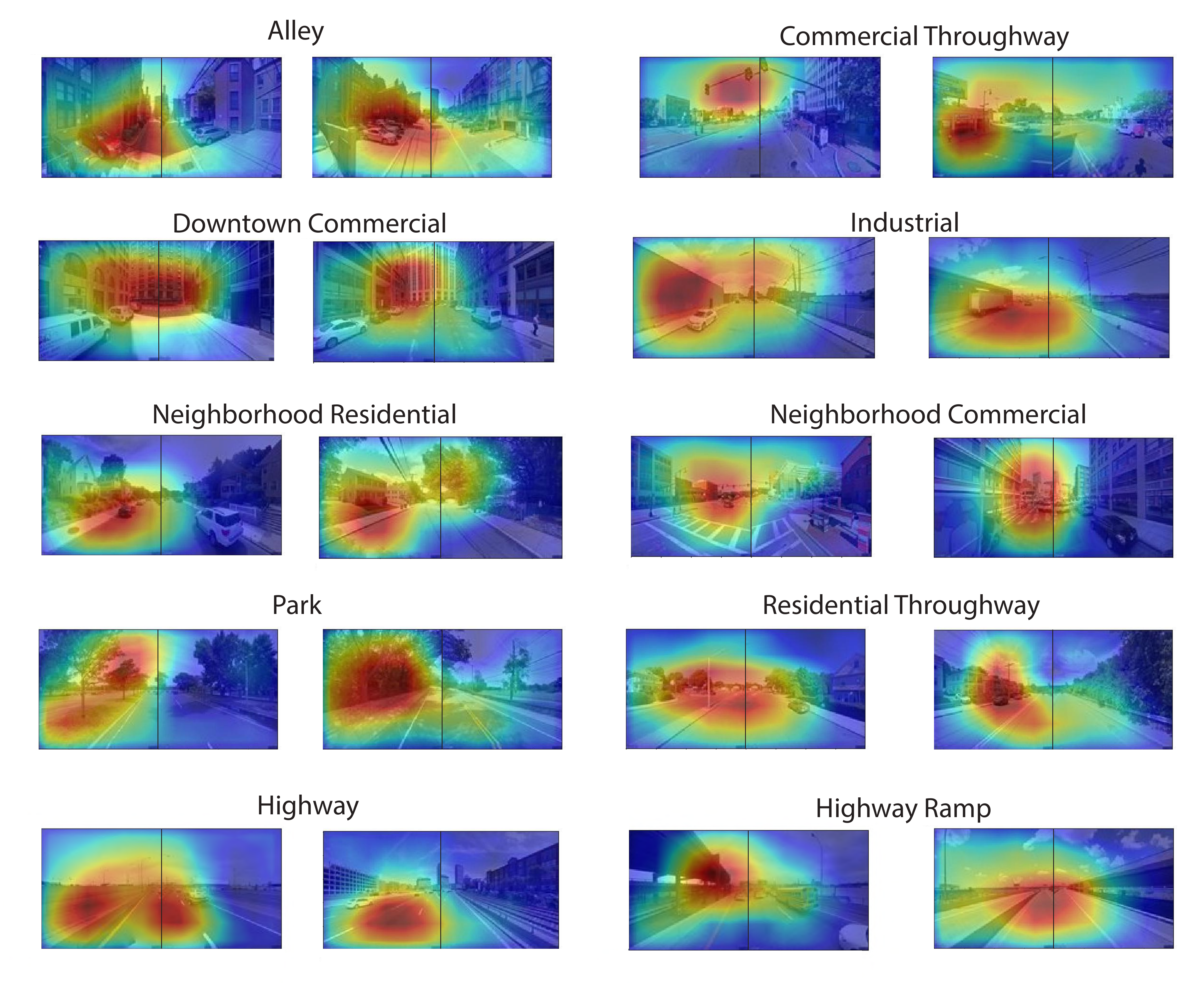}
		\caption{Class activation mapping for street contexts classification in Boston using ResNet18}
		\label{fig:CAM}
	\end{center}
\end{figure*}
Figure \ref{fig:CAM} shows CAM applied to the ResNet18 model for the city of Boston. We show here a sample of the heat maps illustrating some of the features in the street view images that activated certain classification. The model captures several features in images and we discuss some of those shown in Figure \ref{fig:CAM}. For the Alleys class in the figure, the heat map of the CAM highlights red bricked walls, trash bins and back-side parking spots which are features typical for Alleys in the city of Boston.

For Commercial Throughways, we see the activation map highlighting stores on the side of streets, the sidewalk and traffic lights. For Downtown Commercial, we see the activation map hot on high rise buildings typically available in downtown areas. For the Industrial streets, we see the model activated by corrugated steel walls and a cargo truck. For Neighborhood Residential, we see the model was activated by the presence of houses on the side of the street as well as greenery and cars parked on the side. For Neighborhood Commercial, we see the model activated by buildings with stores in them, we also notice the zebra line on the street under the heated activation area. For Parks, the model is activated by trees on the sides of the street. For Residential Throughways, the heated area is often wide to capture the throughway nature of the streets as well as detecting residential houses on the sides. For Highways, the heatmaps highlights the road area of the image on both lanes if exists.  For Highway Ramps, the model is looking at the road ahead which has less number of lanes usually. It is also slightly activated by the presence of a nighboring highway or bridge on which it will merge. These are sample activation features among others that the ResNet18 model highlights towards activating their respective predicted classes. In the Figure \ref{fig:CAM} we show a sample of two images per class for illustration and validation. 

\section{Acknowledgments}
The authors would like to thank the city councils of San Francisco for the street context labeled data. The author would also like to thank the collaboration program (CCES) between MIT and KACST for their continuous support. In addition, the author would like to thank the following individuals, Mohamed Alhajri for valuable inputs, Abdulaziz Alhassan and Abdulaziz Aldawood for interesting discussions and critique.

\bibliographystyle{plain}
\bibliography{citation.bib}

\begin{thebibliography}{10}

\bibitem{streetViewApi}
Google street view api.
\newblock {\url{https://developers.google.com/
  maps/documentation/streetview/intro}}.

\bibitem{Aich_2018_CVPR_Workshops}
Shubhra Aich, William van~der Kamp, and Ian Stavness.
\newblock Semantic binary segmentation using convolutional networks without
  decoders.
\newblock In {\em The IEEE Conference on Computer Vision and Pattern
  Recognition (CVPR) Workshops}, June 2018.

\bibitem{albert2017using}
Adrian Albert, Jasleen Kaur, and Marta~C Gonzalez.
\newblock Using convolutional networks and satellite imagery to identify
  patterns in urban environments at a large scale.
\newblock In {\em Proceedings of the 23rd ACM SIGKDD international conference
  on knowledge discovery and data mining}, pages 1357--1366. ACM, 2017.

\bibitem{albert2017modeling}
Adrian~Toni Albert, E~Strano, and M~Gonzalez.
\newblock Modeling urbanization patterns at a global scale with generative
  adversarial networks.
\newblock In {\em AGU Fall Meeting Abstracts}, 2017.

\bibitem{Buslaev_2018_CVPR_Workshops}
Alexander Buslaev, Selim Seferbekov, Vladimir Iglovikov, and Alexey Shvets.
\newblock Fully convolutional network for automatic road extraction from
  satellite imagery.
\newblock In {\em The IEEE Conference on Computer Vision and Pattern
  Recognition (CVPR) Workshops}, June 2018.

\bibitem{cao2018integrating}
Rui Cao, Jiasong Zhu, Wei Tu, Qingquan Li, Jinzhou Cao, Bozhi Liu, Qian Zhang,
  and Guoping Qiu.
\newblock Integrating aerial and street view images for urban land use
  classification.
\newblock {\em Remote Sensing}, 10(10):1553, 2018.

\bibitem{demir2018deepglobe}
Ilke Demir, Krzysztof Koperski, David Lindenbaum, Guan Pang, Jing Huang, Saikat
  Basu, Forest Hughes, Devis Tuia, and Ramesh Raskar.
\newblock Deepglobe 2018: A challenge to parse the earth through satellite
  images.
\newblock {\em ArXiv e-prints}, 2018.

\bibitem{gebru2017using}
Timnit Gebru, Jonathan Krause, Yilun Wang, Duyun Chen, Jia Deng, Erez~Lieberman
  Aiden, and Li~Fei-Fei.
\newblock Using deep learning and google street view to estimate the
  demographic makeup of neighborhoods across the united states.
\newblock {\em Proceedings of the National Academy of Sciences}, page
  201700035, 2017.

\bibitem{Hamaguchi_2018_CVPR_Workshops}
Ryuhei Hamaguchi and Shuhei Hikosaka.
\newblock Building detection from satellite imagery using ensemble of
  size-specific detectors.
\newblock In {\em The IEEE Conference on Computer Vision and Pattern
  Recognition (CVPR) Workshops}, June 2018.

\bibitem{hamaguchi2018building}
Ryuhei Hamaguchi and Shuhei Hikosaka.
\newblock Building detection from satellite imagery using ensemble of
  size-specific detectors.
\newblock In {\em 2018 IEEE/CVF Conference on Computer Vision and Pattern
  Recognition Workshops (CVPRW)}, pages 223--2234. IEEE, 2018.

\bibitem{he2016deep}
Kaiming He, Xiangyu Zhang, Shaoqing Ren, and Jian Sun.
\newblock Deep residual learning for image recognition.
\newblock In {\em Proceedings of the IEEE conference on computer vision and
  pattern recognition}, pages 770--778, 2016.

\bibitem{kita2019google}
Kinga Kita and {\L}ukasz Kidzi{\'n}ski.
\newblock Google street view image of a house predicts car accident risk of its
  resident.
\newblock {\em arXiv preprint arXiv:1904.05270}, 2019.

\bibitem{krizhevsky2012imagenet}
Alex Krizhevsky, Ilya Sutskever, and Geoffrey~E Hinton.
\newblock Imagenet classification with deep convolutional neural networks.
\newblock In {\em Advances in neural information processing systems}, pages
  1097--1105, 2012.

\bibitem{kuo2018deep}
Tzu-Sheng Kuo, Keng-Sen Tseng, Jia-Wei Yan, Yen-Cheng Liu, and Yu-Chiang~Frank
  Wang.
\newblock Deep aggregation net for land cover classification.
\newblock In {\em CVPR Workshops}, pages 252--256, 2018.

\bibitem{maaten2008visualizing}
Laurens van~der Maaten and Geoffrey Hinton.
\newblock Visualizing data using t-sne.
\newblock {\em Journal of Machine Learning Research}, 9(Nov):2579--2605, 2008.

\bibitem{naik2017computer}
Nikhil Naik, Scott~Duke Kominers, Ramesh Raskar, Edward~L Glaeser, and
  C{\'e}sar~A Hidalgo.
\newblock Computer vision uncovers predictors of physical urban change.
\newblock {\em Proceedings of the National Academy of Sciences},
  114(29):7571--7576, 2017.

\bibitem{naik2014streetscore}
Nikhil Naik, Jade Philipoom, Ramesh Raskar, and C{\'e}sar Hidalgo.
\newblock Streetscore-predicting the perceived safety of one million
  streetscapes.
\newblock In {\em Proceedings of the IEEE Conference on Computer Vision and
  Pattern Recognition Workshops}, pages 779--785, 2014.

\bibitem{SF_BSP}
San~Francisco Planning.
\newblock Better streets plan.
\newblock In {\em https://sfplanning.org/resource/better-streets-plan}, 2010.

\bibitem{sun2018stacked}
Tao Sun, Zehui Chen, Wenxiang Yang, and Yin Wang.
\newblock Stacked u-nets with multi-output for road extraction.
\newblock In {\em 2018 IEEE/CVF Conference on Computer Vision and Pattern
  Recognition Workshops (CVPRW)}, pages 187--1874. IEEE, 2018.

\bibitem{szegedy2016rethinking}
Christian Szegedy, Vincent Vanhoucke, Sergey Ioffe, Jon Shlens, and Zbigniew
  Wojna.
\newblock Rethinking the inception architecture for computer vision.
\newblock In {\em Proceedings of the IEEE conference on computer vision and
  pattern recognition}, pages 2818--2826, 2016.

\bibitem{zhou2016learning}
Bolei Zhou, Aditya Khosla, Agata Lapedriza, Aude Oliva, and Antonio Torralba.
\newblock Learning deep features for discriminative localization.
\newblock In {\em Proceedings of the IEEE conference on computer vision and
  pattern recognition}, pages 2921--2929, 2016.

\bibitem{Zhou_2018_CVPR_Workshops}
Lichen Zhou, Chuang Zhang, and Ming Wu.
\newblock D-linknet: Linknet with pretrained encoder and dilated convolution
  for high resolution satellite imagery road extraction.
\newblock In {\em The IEEE Conference on Computer Vision and Pattern
  Recognition (CVPR) Workshops}, June 2018.

\bibitem{zhou2018d}
Lichen Zhou, Chuang Zhang, and Ming Wu.
\newblock D-linknet: Linknet with pretrained encoder and dilated convolution
  for high resolution satellite imagery road extraction.
\newblock In {\em Proceedings of the IEEE Conference on Computer Vision and
  Pattern Recognition Workshops}, pages 182--186. IEEE, 2018.

\end{thebibliography}

\end{document}